\RequirePackage{fix-cm}
\documentclass{svjour3}                     
\smartqed  
\usepackage{graphicx}
\usepackage{diagbox}
\usepackage{cite}
\usepackage{algorithm}
\usepackage{amsmath}
\usepackage{amsfonts}
\usepackage{color}
\usepackage[noend]{algpseudocode}
\begin{document}

\title{Federated Two Stage Decoupling With Adaptive Personalization Layers}


\author{Hangyu Zhu\and
        Yuxiang Fan\and
        Zhenping Xie
}


\institute{
    H. Zhu \email{hangyu.zhu@jiangnan.edu.cn} \\
    Y. Fan \email{fanyx@stu.jiangnan.edu.cn} \\
    Z. Xie (\texttt{Corresponding author}) \email{xiezp@jiangnan.edu.cn} \\
    School of Artificial Intelligence and Computer Science, Jiangnan University, 1800 Lihu Road, Wuxi 214126, Jiangsu Province, China. 
}

\date{Accepted: 29 Dec 2023(Under production)}

\maketitle

\begin{abstract}
Federated learning has gained significant attention due to its groundbreaking ability to enable distributed learning while maintaining privacy constraints. However, as a consequence of data heterogeneity among decentralized devices, it inherently experiences significant learning degradation and slow convergence speed. Therefore, it is natural to employ the concept of clustering homogeneous clients into the same group, allowing only the model weights within each group to be aggregated. While most existing clustered federated learning methods employ either model gradients or inference outputs as metrics for client partitioning to group similar devices together, heterogeneity may still exist within each cluster. Moreover, there is a scarcity of research exploring the underlying reasons for determining the appropriate timing for clustering, resulting in the common practice of assigning each client to its own individual cluster, particularly in the context of highly non independent and identically distributed (Non-IID) data. In this paper, we introduce a two-stage decoupling federated learning algorithm with adaptive personalization layers named FedTSDP, where client clustering is performed twice according to inference outputs and model weights, respectively. Hopkins amended sampling is adopted to determine the appropriate timing for clustering and the sampling weight of public unlabeled data. In addition, a simple yet effective approach is developed to adaptively adjust the personalization layers based on varying degrees of data skew. Experimental results show that our proposed method has reliable performance on both IID and non-IID scenarios.
\end{abstract}

\section{Introduction}
Federated learning (FL) \cite{fedavg} is an emerging privacy-preserving machine learning scheme. It allows multiple participants to collaboratively learn a shared global model without sharing private local data, thus, effectively alleviating the barrier of data silos. And distributed devices with high privacy concerns can benefit from this learning process. Nowadays, FL has already been widely adopted in many real-world scenarios, such as disease diagnosis\cite{chest},  edge computing\cite{edge}, autopilot\cite{autodrive}, and so on. While vanilla FL approaches are not robust to non independent and identically distributed (Non-IID) data, and enormous research works \cite{zhao2018federated,ZHU2021371} have indicated that performance deterioration of a single global model in FL is almost inevitable on Non-IID or heterogeneous data. Although local models are often initialized with the same global model at each communication round, they would converge to different directions due to data heterogeneity. Consequently, the divergence between the global model (averaged by local models) and the ideal model accumulates over the training period, significantly slowing down the convergence speed and worsening the learning performance.

To address the divergence issue mentioned above, a plethora of methodologies have been proposed in the literature \cite{9155494,8889996,li2021fedbn}. Data sharing \cite{zhao2018federated,9412599,TIAN2021102344} represents a straightforward yet efficacious approach aimed at alleviating the adverse impact of non-IID data. The fundamental concept entails priming the global model by incorporating a fraction of the shared dataset on each client, thereby mitigating the aforementioned side effects. However, it is important to note that this approach may introduce additional local data leakage, potentially compromising the privacy requirements of federated learning to some degree. Regular optimization \cite{MA2022244,9183470} serves as another prevalent approach for addressing non-IID challenges by incorporating an auxiliary regularization term into the local loss function. For instance, the FedProx algorithm \cite{fedprox} employs the $l_2$ norm of the distance between the global and local models as the regularization term. By optimizing the local loss function with this regularization term, the local updates are guided to be more aligned with the global model, thereby mitigating model divergence arising from data heterogeneity. Similar principles are also employed in Ditto \cite{ditto}, wherein the regularization term is selectively introduced after a predetermined number of FL rounds. One shortcoming of regular optimization approaches is that they always consume more computational resources than the conventional FedAvg \cite{fedavg}. Meanwhile, certain literature \cite{9835537} indicates that the regularization term might not provide benefits in the presence of data heterogeneity, particularly in highly non-IID scenarios.

Inspired by the principles of multi-task \cite{multitask} and transfer learning \cite{weiss2016survey}, the inclusion of personalization layers in FL has been proposed as a means to solve highly non-IID problems. Arivazhagan \emph{et al.} introduced FedPer \cite{fedper}, which employs shallow base layers for high-level representation extraction and deep personalization layers for classification. As the personalization layers in FedPer are not shared or aggregated on the server, the local model's statistical characteristics are retained. And FedRep \cite{fedrep}, an extension of FedPer, focuses on extracting representations specifically from personalized layers. By contrast, LG-FedAvg \cite{liang2020think} incorporates shallow layers as personalization layers and deep layers as base layers. FedAlt \cite{fedalt} conducted a comprehensive investigation, including convergence analysis, to explore the influence and architectural considerations related to personalized layers. And it is found that the last few personalization layers have a direct impact on learning bias and serve as a key contributing factor to model divergence \cite{proveforclassifier}. In a more recent development, Tashakori \emph{et al.} proposed the SemiPFL framework \cite{semi} as a novel approach to amalgamate personalized federated learning and semi-supervised learning, with the objective of enhancing multi-sensory classifications.

The aforementioned approaches, by default, employ the aggregation of all uploaded model parameters from connected clients. However, this aggregation method is less advantageous for Non-IID data scenarios due to the inherent limitations of a single global model in effectively accommodating all local learning tasks. Therefore, clustered FL presents itself as a promising solution in such contexts, where clients exhibiting similarity in their models are assigned to the same cluster group, and only models within the same cluster are aggregated. CFL\cite{cfl}, regarded as the pioneering clustered FL algorithm, employs gradient information to recursively bi-partition clients, effectively mitigating gradient conflicts. Nevertheless, in high-dimensional spaces, gradient information can often be ambiguous, leading to potential variations in convergence even among clients with the same bias. IFCA \cite{ifca} represents another noteworthy approach in the realm of clustered FL. And it leverages loss information to estimate cluster identities, effectively grouping together well-trained clients within the same cluster. Over the past three years, an increasing number of clustered FL methods have emerged. These methods leverage various criteria, including gradient information \cite{flhc,cic,flex,rcfl}, model weights \cite{fedsoft,fedsem,stratified}, and auxiliary models \cite{comet,flis}, to identify clusters under different constraints, such as economic considerations, efficiency requirements, and low latency constraints \cite{auction,10061474,smarthome,comm-efficient,fedtcr}. Yet, all the clients can be easily allocated to distinct clusters for highly heterogeneous local data, where each cluster comprises a single client, rendering collaborative training no longer meaningful. Hence, the integration of cluster FL and personalized FL becomes imperative in order to mitigate this issue. And to the best of our knowledge, there exists limited research pertaining to this particular aspect.

Therefore, in this work, we propose a novel two-stage decoupling personalized FL algorithm called \textbf{FedTSDP}, where client models with dynamic personalization layers are clustered twice based on model inference and weights, respectively. The main contributions of this paper are listed as follows:

\begin{enumerate}
    \item This is the first work that extends the conventional single-stage clustered FL to a more sophisticated two-stage scheme. The first stage leverages the outcomes of model inference to gauge the preference of different clients, while the second stage utilizes model weights to assess their respective local learning directions, thus, exhibiting faster convergence speed compared to the single-stage clustering method.
    \item For the first stage of clustering, the Jensen-Shannon (JS) divergence \cite{nielsen2021variational} is employed as a similarity metric to quantify the divergence among participating clients. Furthermore, a Hopkins \cite{hopkins} amended sampling method is introduced to simultaneously determine the appropriate timing for clustering and ascertain the sampling weight of public unlabeled data on the server.
    \item For the second stage, model weights are utilized as metrics to conduct clustering within each group formed during the first stage. In addition, a simple yet effective paradigm is developed to dynamically adjust the number of personalization layers according to varying degrees of data skew in FL.
    \item Empirical experiments are conducted to showcase the promising performance achieved by the proposed FedTSDP algorithm on both heterogeneous and homogeneous client data.
\end{enumerate}

\section{Background and Motivation}
In this section, a concise overview of federated learning with personalization layers is given at first, followed by an introduction to clustered federated learning. Lastly, the motivation of the present work is reiterated.

\subsection{Federated Learning with Personalization Layers}
The original FL algorithm, known as FedAvg \cite{fedavg}, aims to find the optimal global server model $w$ capable of minimize total $K$ aggregated local loss $F_{i}(w_{i})$ of each client $i$ as shown in Eq. \eqref{eq:fedavg}: 

\begin{equation}
\label{eq:fedavg}
F_{i}(w_{i})=l(w_{i}; \mathcal{D}_{i});  \quad  \min_w{f(w)} = \sum_{i=1}^{K}{\frac{n_{i}}{n} F_i(w_{i})},
\end{equation}
where $\mathcal{D}_{i}$ is the local training data, $w_{i}$ is the model weights or parameters. And $n_{i}=\left| \mathcal{D}_{i} \right|$ is the local data size, $n=\sum_{i=1}^{K}n_{i}$ is the total data size, respectively. In FedAvg, a uniform weight initialization is applied to all local models sharing the same architecture at each communication round. Subsequently, each client returns its entire model weights to the server for \emph{federated aggregation} (right panel of Eq. \eqref{eq:fedavg}) after local training.

\begin{figure}
\centering 
\includegraphics[width=0.92\textwidth]{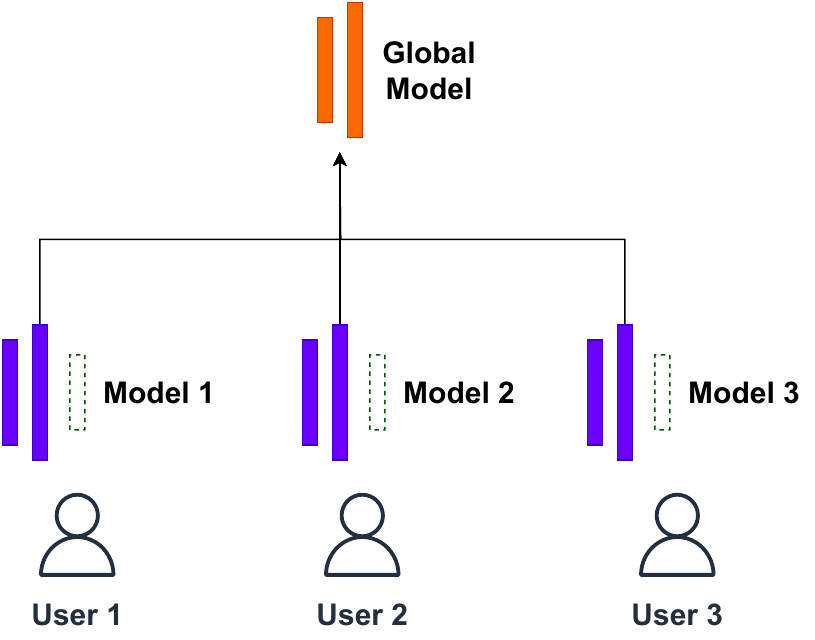} 
\caption{A simple example of FL with personalization layers. The filled blocks of each user are shallow layers, while the dashed blocks are personalization layers.}  
\label{fig:fedper}
\end{figure}

However, as a result of data heterogeneity, direct federated aggregation upon the entire client models may lead to model divergence. \cite{zhao2018federated}. Hence, it is imperative to introduce the concept of FL with personalization layers \cite{fedper} as a remedy for the aforementioned issue. As depicted in Fig. \ref{fig:fedper}, where the shallow layers of the neural network model are base layers uploaded to the server for aggregation, while deep layers are personalization layers stored locally on each user. During each communication round of the training period, every client receives identical base layers from the server and incorporates them with their respective personalization layers. The resulting integrated local model, which comprises both the base and personalization layers, is then updated through training on local data. After that, only the updated base layers are returned back to the server for subsequent model aggregation. This process not only facilitates collaborative learning but also preserves the personalized features of each user's model.

And several work \cite{fedrep, liang2020think} have been proposed to enhance above mentioned personalized FL scheme by identifying the optimal placement of personalized layers. This is motivated by the inherent capability of personalization layers to encapsulate significant pattern signals, which aids in the identification of participants' bias. As a result, determining the appropriate number of personalization layers becomes a critical factor for the successful implementation of personalized FL. 

\subsection{Clustered Federated Learning}
Another effective approach addressing data heterogeneity issue is to partition the connected clients into several groups or clusters, which particularly relevant in scenarios where distinct user groups possess respective learning objectives. Yet through the aggregation of their local models with others in the same cluster sharing similar tasks, they can harness the power of collective intelligence to achieve more efficient FL.

This is the reason why clustered FL has emerged as a promising solution to Non-IID data. As shown in Fig. \ref{fig:cfl}, each client initiates local training and subsequently uploads the trained local model to the server. The server proceeds to perform client clustering based on metrics, typically computed on each client using Eq. \eqref{eq:ifca}:

\begin{figure}
\centering
\includegraphics[width=0.92\textwidth]{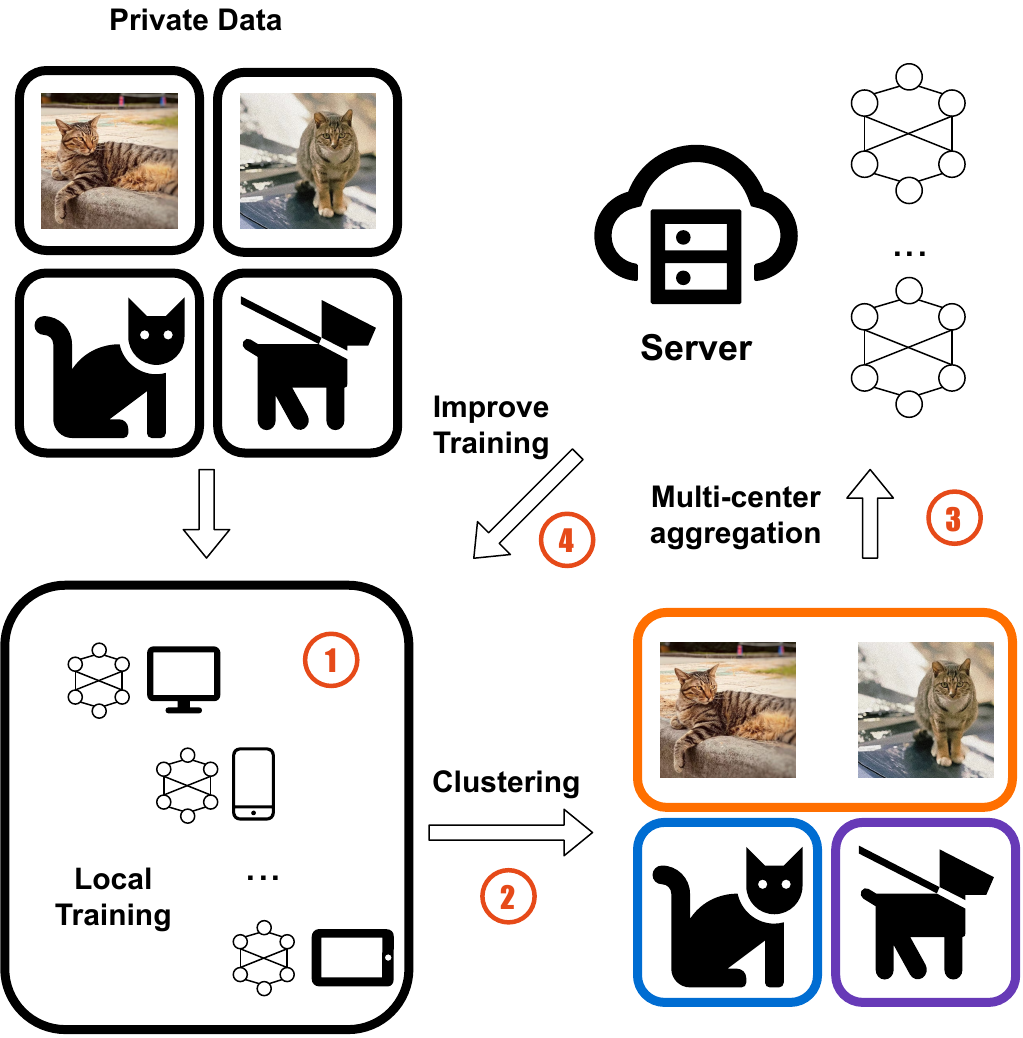}
\caption{An illustrative example of clustered FL (boxes with the same color represent a cluster). \textcircled{\scriptsize{1}} Perform local training on connected clients. \textcircled{\scriptsize{2}} Perform clustering to partition clients into different groups. \textcircled{\scriptsize{3}} Perform multi-center aggregation on the server. \textcircled{\scriptsize{4}} Clients receive the model of the corresponding cluster from the server.}
\label{fig:cfl}
\end{figure}

\begin{equation}
\label{eq:ifca}
\hat{j}=\text{argmin}_{j\in [J]}F_{i}(w_{j}),
\end{equation}
where $j$ is the cluster index, and $\hat{j}$ denotes the cluster index for client $i$ with the lowest local empirical loss $F_{i}(w_{j})$. This is actually the core concept introduced in IFCA algorithm \cite{ifca}, wherein $J$ global models are simultaneously downloaded to connected clients,  and each client is assigned to the group corresponding to the lowest $F_{i}(w_{j})$ value. Other methodologies like CFL \cite{cfl} utilize the distance of gradient information as a metric to reduce both communication costs and local computational costs.

After client clustering, the server conducts multi-center aggregation within each group and subsequently distributes the aggregated global models to the respective clients belonging to the corresponding cluster. Note that, in most clustered FL algorithms, the number of clusters, denoted as $k$, is predetermined and remains constant throughout the training process.

However, it is likely that each client is assigned to its own individual cluster, resulting in one client per cluster, renders the clustering process meaningless, particularly in highly Non-IID data scenarios. Consequently, the integration of clustered FL and personalized FL becomes imperative to address this issue and limited research has been conducted in this specific area to date.

\section{Proposed Algorithm}
In this section, the proposed FedTSDP algorithm would be discussed in detail. At first, We would like to introduce the methodology of the first stage decoupling including the calculation of similarity metrics based on Jensen-Shannon (JS) divergence and Hopkins amended sampling. Subsequently, weight constraint based clustering together with adaptive personalization layer adjustment for the second stage are presented, aiming to further reduce the client model divergence. And the overall framework of FedTSDP is illustrated at the last.

\subsection{Problem Description}
In this work, we consider a setting of clustered FL with personalization layers where each user works as a client and the server possesses an unlabelled dataset. Let $\mathcal{C}_{j}^{l}, j=1,\ldots, J, l = 1, 2$ denotes the $j$-th group of the $l$-th stage clustering, $\mathcal{D}_{i}$ and $\mathcal{D}_{u}$ are the training data of client $i$ and public \emph{unlabelled} data on the server, respectively, $\left[ \mathcal{L}^{s};\mathcal{L}^{per} \right]$ represents all layers of the learning model where $\mathcal{L}^{s}$ is the number of shared (base) layers and $\mathcal{L}^{per}$ is the number of personalization layers. It aims to simultaneously find a group of optimal model weights  $\left\{w_{1}^{*},w_{2}^{*}, \ldots, w_{J}^{*}\right\}$ for all clusters and their (appropriate) number of personalized layers $\mathcal{L}^{per}$. And all the other notations used in this section are listed in Table.\ref{tab:notations}.

\begin{table}[ht]
\caption{Notation Lists}
\label{tab:notations}
\begin{tabular}{ll}
    \hline
        Symbol & Description \\
    \hline
        $\mathcal{D}_i$ & Dataset of the $i$-th client \\
        $\mathcal{D}_{pub}$ & Public unlabelled dataset on the server \\
        $\mathcal{C}_{j}^{l}$ & The $j$-th cluster of the $l$-th stage \\
        $W_{k}^{r}$ & The sampling weight of data point $k$ at round $r$ \\
        $\epsilon$ & Epsilon threshold for DBSCAN \\
        $\text{min}_{pts}$ & Minimum Points threshold for DBSCAN \\
        $H$ & Hopkins statistics \\
        $h_{\text{th}}$ & Hopkins statistics threshold \\
        $\mathcal{L}_{j}^{s}$ & The number of shared layers of the $j$-th cluster  \\
        $\mathcal{L}_{j}^{per}$  & The number of personalization layers of the $j$-th cluster \\
        $F_{i}$ & The learning objective of client $i$ \\
        $E$ & Number of local epochs \\
        $I$ & A set of inference outputs \\
        $\psi^{r}$ & Dampening ratio of the shared layers. \\
    \hline
\end{tabular}
\end{table}

\subsection{The First Stage of Decoupling}
Unlike CFL \cite{cfl} adopting the distance of model gradients to recursively bi-partition clients, the first stage of our proposed method employs the inference outputs of the learning model from each connected client for clustering. The underlying reason is that the inference output inherently encapsulates the characteristics of data information and possesses better ability to reflect the local data distribution compared to model gradients.

\begin{figure}[ht]
  \includegraphics[width=0.92\textwidth]{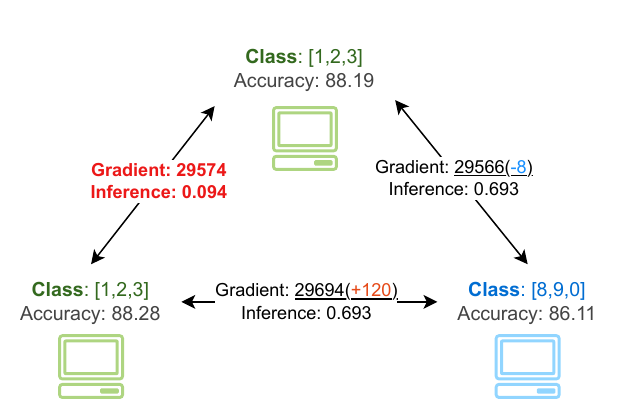}
\caption{A simple FL example to demonstrate inadequacy of using model gradients as a clustering metric, where clients with same color indicate that they possess training data with the same label classes.}
\label{fig:sub}
\end{figure}

As a simple FL example shown in Fig. \ref{fig:sub}, wherein three different clients possesses training data with label classes $\left[ 1,2,3 \right]$, $\left[ 1,2,3 \right]$, and $\left[ 8,9,0 \right]$, respectively. The learning model is trained using FedAvg until convergence, after which two types of distances between clients are evaluated based on the model gradients and inference outputs in pairwise comparisons. It is evident that the client belonging to the label class $\left[8, 9, 0\right]$ exhibits a notable difference in gradient distance to the other two clients belonging to $\left[1, 2, 3\right]$ (one is 29694 and the other is 29566), whereas the inference distance demonstrates a robust relationship with the underlying data distribution. This empirical evidence suggests that inference outputs have a stronger capability to represent the local data distribution compared to gradient information, thus, is proper to be the clustering metrics. And in the subsequent section, we will introduce the methodology for calculating the distance based on the inference outputs.

\subsubsection{Jensen-Shannon Divergence}
\label{sub:js}
Aforementioned inference distance can also be interpreted as a similarity measure when considering pairwise comparisons between clients. And Kullback-Leibler (KL) divergence \cite{csiszar1975divergence} is widely used as a technique to quantify the dissimilarity between two probability distributions. 

Given two probability distributions $P$ and $Q$, the KL divergence $D_{\text{KL}}(P||Q)$ from $P$ to $Q$ is defined in the following Eq. \eqref{eq:kl}:

\begin{equation}
\label{eq:kl}
    D_{\text{KL}}(P||Q)=\sum_{x\in \mathcal{X}}P(x)\log{\frac{P(x)}{Q(x)}},
\end{equation}
where $\mathcal{X}$ is the set of all possible outcomes, $P(x)$ is the probability of the outcome $x$ according to distribution $P$, and $Q(x)$ is the probability of the outcome $x$ based on distribution $Q$. For a typical classification problem in machine learning, the model inference is often represented as logits which can be interpreted as a discrete probability distribution. And during the training process of FL, once the server receives model weights $w_{i}$, $w_{i^{'}}$ from two distinct clients, their corresponding inferences can be computed as $p_{i}=F_{i}(w_{i};\mathcal{D}_{pub})$ and $p_{i^{'}}=F_{i^{'}}(w_{i^{'}};\mathcal{D}_{pub})$, respectively, using public unlabelled dataset $\mathcal{D}_{pub}$. Then, the similarity from client $i$ to client $i^{'}$ can be computed using the KL divergence as shown in Eq. \eqref{eq:klc}:

\begin{equation}
\label{eq:klc}
    D_{\text{KL}}(p_{i}||p_{i^{'}})=\sum_{c \in C}p_{i,c}\log{\frac{p_{i,c}}{p_{i^{'},c}}},
\end{equation}
where $c$ is the class index of the model inference, and $p_{i}$ satisfies $\sum_{c}p_{i,c}=1$. However, it is important to note that KL divergence is not symmetric and does not measure the physical distance in Hilbert space, which means $D_{\text{KL}}(p_{i}||p_{i^{'}}) \neq D_{\text{KL}}(p_{i^{'}}||p_{i})$. This property is not suitable for client clustering, as it implies that the pairwise distances between the same two clients may not be equal. Consequently, Jensen-Shannon (JS) Divergence \cite{menendez1997jensen} is instead applied here in Eq. \eqref{eq:js}:

\begin{equation}
\label{eq:js}
    D_{\text{JS}}(p_{i}||p_{i^{'}})=\frac{1}{2}D_{\text{KL}}(p_{i}||\hat{p}_{i,i^{'}})+\frac{1}{2}D_{\text{KL}}(p_{i^{'}}||\hat{p}_{i,i^{'}}),
\end{equation}
where $\hat{p}_{i,i^{'}}=\frac{1}{2}(p_{i}+p_{i^{'}})$ is the average distribution of $p_{i}$ and $p_{i^{'}}$, and it has been theoretically proved that $D_{\text{JS}}(p_{i}||p_{i^{'}})=D_{\text{JS}}(p_{i^{'}}||p_{i})$.

\begin{figure}[ht]
  \includegraphics[width=0.92\textwidth]{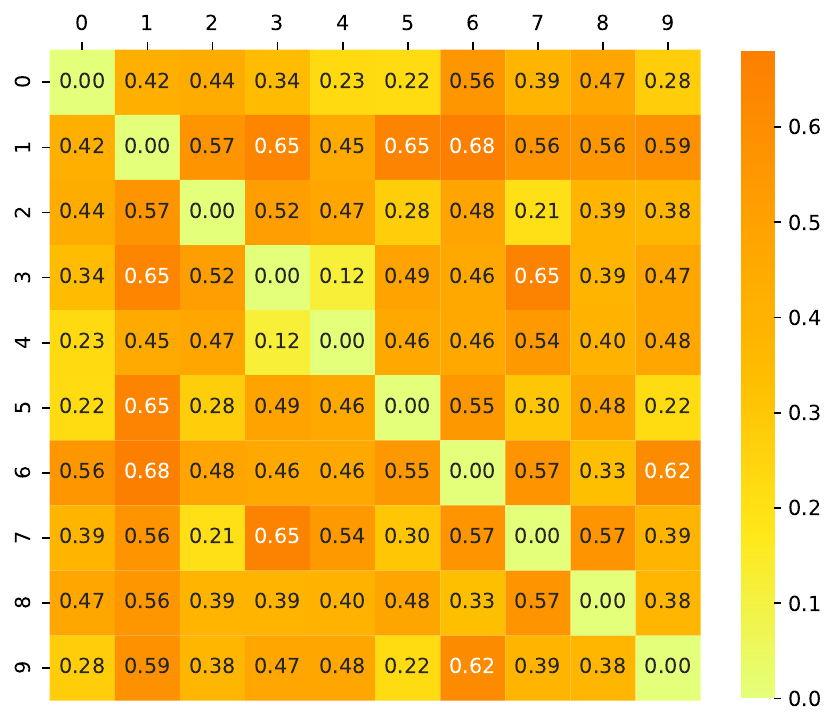}
\caption{An example of the similarity matrix computed among 10 clients, where the value of each element is calculated by Eq. \eqref{eq:js}. The smaller the value, the more similar two clients will be.}
\label{fig:js_map}
\end{figure}

Assuming there are a total of $m$ connected clients in FL system, the server can construct an $m \times m$ similarity matrix, where each element $\text{elem}_{i,i^{'}}$ represents the computed result of JS divergence $D_{\text{JS}}(p_{i}||p_{i^{'}})$ between the inference outputs of two client models $w_{i}$ and $w_{i^{'}}$. An illustrative example of $10 \times 10$ similarity matrix is depicted in Fig. \ref{fig:js_map}, where both horizontal and vertical axes represent client indices. The values within each square of the matrix, calculated using JS divergence, indicate the similarity between two clients corresponding to the respective client indices. It is evident that the smaller the value, the greater the similarity between two clients. And this similarity matrix serves as the metric for the subsequent clustering algorithm. In our proposed scheme, we select density-based spatial clustering of applications with noise (DBSCAN) \cite{ester1996density} as the clustering algorithm due to its ability to determine the number of clusters without prior specification.

\subsubsection{Hopkins Amended Sampling}
\label{sec:hopkins}
As discussed previously, the timing of clustering plays a crucial role in clustered FL algorithms. For those scenarios of high Non-IID data, it is likely that each client is assigned to its own cluster. And this actually is equivalent to single client training, rendering collaborative learning meaningless.

To address above mentioned issue, Hopkins statistic is adopted to automatically specify the timing of clustering. The Hopkins statistic, originally a quantitative measure used to assess the clustering tendency of a dataset, is adapted in our FL scenarios to evaluate the clustering tendency based on inference outputs of client models. Let $I = \{p_{i} \in \mathbb{R}^{d} | i=1...m \}$ denote a set of $m$ \emph{flattened} inferences where the vector dimension $d=B \times C$. $B$ is the batch size of the public data, and $C$ is the total number of classification outputs. Note that, the reason for adopting only batch data instead of the entire public dataset for inference computation is to reduce the computational overhead. And also given $\widetilde{I}$ be a set of $\widetilde{m} \ll m$ elements randomly sampled without replacement from $I$, $\widehat{I}$ be a set of $\widetilde{m}$ elements uniformly randomly sampled from the sampling space of $I$. Then, the Hopkins statistic can be defined as Eq. \eqref{eq:hopkins}:

\begin{equation}
\label{eq:hopkins}
    H=\frac{\sum^{\widetilde{m}}_{i=1}{z_{i}}}{\sum^{\widetilde{m}}_{i=1}{z_i}+\sum^{\widetilde{m}}_{i=1}{v_i}},
\end{equation}
where $z_{i} \in \mathbb{R}$ is the minimum distance of $\widehat{p}_{i} \in \widehat{I}$ to its nearest neighbour in $I$, and $v_{i} \in \mathbb{R}$ is the minimum distance of $\widetilde{p}_{i} \in \widetilde{I} \subseteq I$ to its nearest neighbour $p_{i^{'}} \in I$, where $\widetilde{p}_{i} \neq p_{i^{'}}$. The L2-norm is selected as the metric for computing the pairwise distances.

The resulting Hopkins statistic $H$ is compared with a predefined threshold $h_\text{th}$ afterwards. If $H$ is greater than $h_\text{th}$, client clustering is performed using DBSCAN algorithm, and vice versa. Furthermore, the sampling weights of the aforementioned batch data would be updated to facilitate successful clustering operations as shown in Eq. \eqref{eq:sample}:

\begin{equation}
\label{eq:sample}
     W_{k}^{r-1}=W_{k}^{r-1}+\frac{|\mathcal{D}_{pub}|}{|\mathcal{D}_B|}, k \in \left[ B \right],
\end{equation}
where $W_{k}^{r-1}$ is the sampling weight or ratio of data $k$ at communication round $r-1$, $|\mathcal{D}_{pub}|$ indicates the total size of the public unlabelled dataset, $|\mathcal{D}_B|$ represents the batch size, and $\mathcal{D}_B \subseteq \mathcal{D}_{pub}, |\mathcal{D}_B| \ll  |\mathcal{D}_{pub}|$. It is important to note that only the sampling weights of data within the corresponding batch data $\mathcal{D}_B$ are allowed to be updated. Afterwards, the new sampling weights $W_{k}^{r}$ for the next communication round are obtained by normalizing the updated weights $ W_{k}^{r-1}$, as shown in Eq. \eqref{eq:sample1}:

\begin{equation}
\label{eq:sample1}
    W_{k}^{r}=\frac{W_{k}^{r-1}}{\sum_{k=1}^{|\mathcal{D}_{pub}|}W_{k}^{r-1}}.
\end{equation}

The objective of updating the sampling weights is to increase the likelihood of sampling public data with inference outputs that satisfy the Hopkins statistic. In other words, the sampling process prioritizes data samples that exhibit substantial discrepancies among different client models, thereby significantly enhancing the clustering tendency in the first stage.

\subsection{The Second Stage of Decoupling}
The second stage of decoupling builds upon the results obtained from the first stage, further refining the clustering process. And its initial purpose is to further minimize the distance of model weights between clients in pairwise comparisons. Except that, personalization layers are introduced as an additional component, which not only provides novel assistance to the clustering process but also aims to preserve the statistical characteristics of local models, particularly in the context of Non-IID data.

\subsubsection{Weight Constraint based Clustering}
In this stage, the model weights of each client are used as the clustering metric instead, since they directly represent the properties of the models, unlike gradients or other auxiliary information. And the pairwise distance between any two clients are calculated by Eq. \eqref{eq:modeldist}:

\begin{equation}
\label{eq:modeldist}
    \text{dist}(w_{i},w_{i^{'}})=\left\| w_{i}-w_{i^{'}}+\Upsilon e \right\|_{2},
\end{equation}
where $w_{i}$ and $w_{i^{'}}$ are model weights of client $i$ and client $i^{'}$, respectively, $\Upsilon$ is a small constant added to avoid division by zero, $e$ represents the vector of ones. $\left\|*\right\|_{2}$ is the L2-norm given by Eq. \eqref{eq:l2norm}:

\begin{equation}
\label{eq:l2norm}
    \left\| w \right\|_{2}=\left( \sum_{i=1}^{m}\left| w_{i} \right|^{2} \right)^{1/2}.
\end{equation}

\begin{figure}[ht]
  \includegraphics[width=0.92\textwidth]{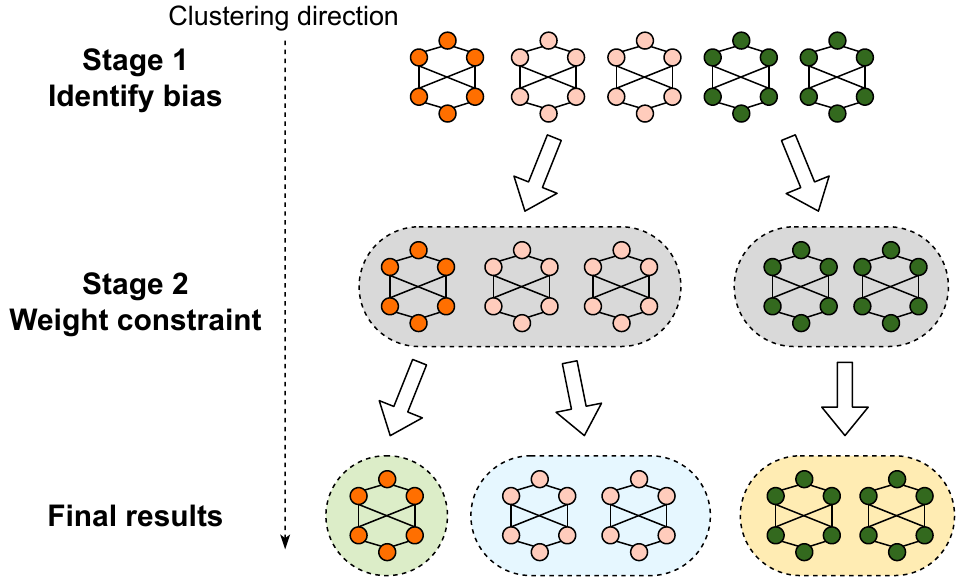}
\caption{An example of two-stage decoupling process, where 5 client models are divided into two groups by the first stage of clustering. And the second stage is performed within each clustered group and continues to, for example, partition 3 client models located at the left panel into two distinct clusters.}
\label{fig:fedtsd}
\end{figure}

As mentioned earlier, this step continues to perform clustering within each group that was formed in the previous stage. The reason for this operation is that the convergence direction of the models may differ, even for data with a homogeneous distribution (shown in Fig. \ref{fig:sub}). These two stages are combined to form a directed hierarchical decoupling flow, as depicted in Fig. \ref{fig:fedtsd}, where 5 client models at the first stage are partitioned into two different groups. And weight constraint based clustering is performed afterwards with each clustered group and continues to partition, for instance, 3 models located at the left panel into two distinct clusters.

\subsubsection{Adaptive Personalization Layer Adjustment}
Despite the ability of our proposed hierarchical decoupling method to partially reduce model divergence resulting from distinct local data distributions, model heterogeneity still persists even after the two-stage decoupling process. Consequently, the integration of personalization layers is introduced as an effective measure to adequately address this concern.

\begin{figure}[ht]
  \includegraphics[width=0.92\textwidth]{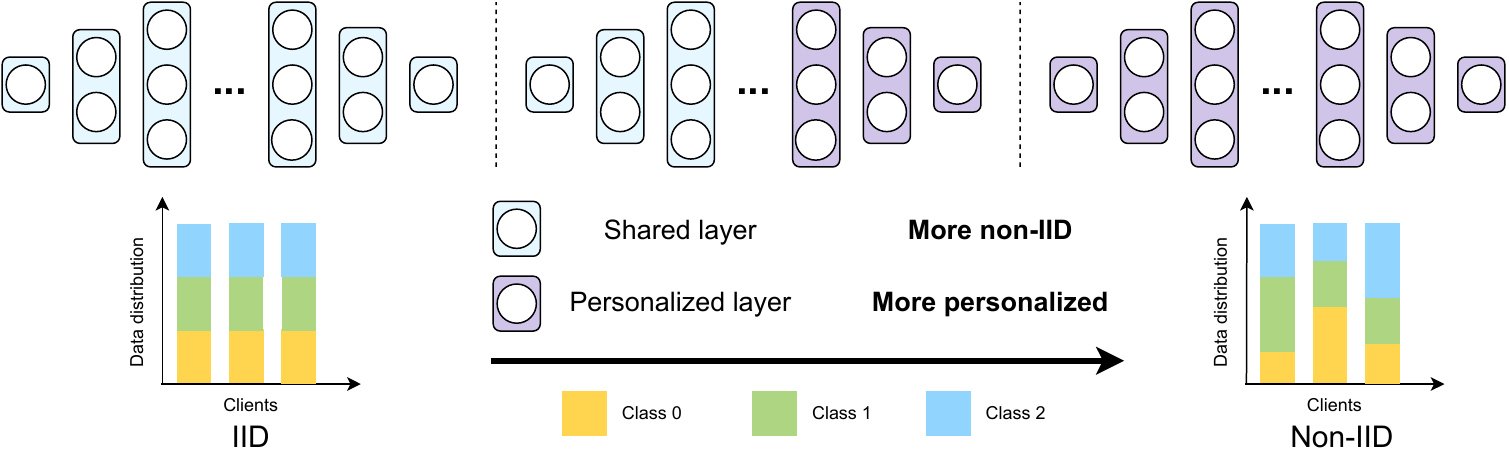}
\caption{An example indicates the relationship between data skew and the number of personalization layers. For more Non-IID data, the number of shared layers should decrease to alleviate the impact of performance degradation, and vice versa. As a result, each local model is able to reserve more individual local attributes, avoiding possible side effect from federated model aggregation caused by Non-IID data.}
\label{fig:dynamic_layer}
\end{figure}

Nevertheless, determining the optimal number of personalization layers necessary to handle varying levels of data skew, as well as their integration with our proposed decoupling scheme, remains uncertain. Furthermore, it is worth noting that many prior research efforts overlook the fact that personalization layers do not always contribute positively to the learning performance of federated learning, especially in scenarios where data follows IID distribution. As illustrated in Fig. \ref{fig:dynamic_layer}, to achieve a relatively similar global model performance, the number of personalization layers exhibits an increasing trend as the client data distribution transitions from IID to non-IID, which means for more Non-IID data, each client is expected to reserve more personalization layers without being shared to the server to preserve more local data attributes, and vice versa.

This observation is reasonable, as aggregating models with all layers trained on significantly different datasets would be harmful and result in significant performance degradation. Conversely, aggregating models with only a few layers trained on similar datasets would lead to a reduced learning capacity, as they would capture less diverse information. Hence, it would be beneficial to dynamically adjust the number of personalization (shared) layers, allowing the learning model to automatically adapt to varying degrees of data skew. However, due to the strict prohibition on accessing local training data in FL, conducting the aforementioned layer adjustment becomes challenging.

Naturally, a connection can be established between data skew and the application of Hopkins statistic theory in the first stage of decoupling. The degrees of data heterogeneity can be quantitatively assessed through clustering tendency analysis. Specifically, a higher clustering tendency indicates greater data divergence, while a lower clustering tendency suggests lower data divergence. And our proposed layer adjustment scheme is performed by decaying the number of shared layers $\mathcal{L}^{s}$ (equivalent to increasing the number of personalization layers $\mathcal{L}^{per}$) as long as Hopkins statistic criterion $H$ is satisfied. As shown in Eq. \eqref{eq:layer_decay}:

\begin{equation}
\label{eq:layer_decay}
    \mathcal{L}^{s} = \mathcal{L}^{s} \times \psi^{r},
\end{equation}
where $\psi^{r}$ is the dampening ratio at the communication round $r$.

\subsection{Overall Framework}
Both clustering and personalization layers are integrated to form a federated two-stage decoupling with adaptive personalization layers approach named FedTSDP. The complete pseudo code illustrating the overall procedure is presented in Algorithm.\ref{algotirhm:fedtsd}.

\begin{algorithm}
\caption{Federated Two-Stage Decoupling With Adaptive Personalization Layers (FedTSDP)}\label{algotirhm:fedtsd}
\begin{algorithmic}[1]
 \State \textbf{Server executes: }
 \State Initialize the global model $w$
 \State Let $\mathcal{C}^{2} = \emptyset$
 \State Initialize $\left[ \mathcal{L}_{s};\mathcal{L}_{per} \right]$
    \For{each round $r=1,2,\ldots,R$}
            \State $m \gets \text{max}\left( \delta \cdot K, 1 \right)$
            \State $\mathcal{C} \gets \left\{ 1,2,...,m \right\}$
            \If{$\mathcal{C}^{2} = \emptyset$}
            \State $\mathcal{C}^{2} \gets \mathcal{C}$
            \EndIf
            \State Send $w_{\mathcal{L}_{j}^{s}}^{r}$ to client $i \in \left[ m \right]$ belonging to cluster $j$
            \For{each client $i=1,2,\ldots, m$ \textbf{in parallel}}
                \State \textbf{Client} $i$ \textbf{executes} local update and returns $w_{i}$
                \State Sample $\mathcal{D}^{B}_{pub} \in \mathcal{D}_{pub}$
                \State $p_{i} \gets F(w_{i};\mathcal{D}_{pub}^{B})$ 
            \EndFor
            \State $I \gets \left\{ p_{i} | i \in [m] \right\}$
            \State Calculate $H$ using $I$ (Eq. \eqref{eq:hopkins})
            \If{$H > h_{\text{th}}$}
            \State // \textbf{\emph{The first stage of decoupling}}
            \State Calculate $\text{Sim}_{i,i^{'}} \gets D_{\text{JS}}(p_{i}||p_{i^{'}}), i, i^{'} \in [m]$ (Eq. \eqref{eq:js})
            \State Perform clustering $\mathcal{C}^{1} \gets \text{DBSCAN}(\mathcal{C}, \text{Sim}, \epsilon_{1}, \text{min}_{pts})$
            \For{each $\mathcal{C}^{1}_{j}, j= 1,2,...,J$ }
            \State // \textbf{\emph{The second stage of decoupling}}
            \State Calculate $D_{i,i^{'}} \gets \text{dist}(w_{i},w_{i^{'}}), i, i^{'} \in \left[  \mathcal{C}^{1}_{j} \right]$ (Eq. \eqref{eq:modeldist})
            \State Perform clustering $\mathcal{C}^{2}_{j} \gets \text{DBSCAN}(\mathcal{C}^{1}_{j}, D, \epsilon_{2}, \text{min}_{pts})$
            \EndFor
            \State Update $\mathcal{C}^{2} \gets \left\{ \mathcal{C}^{2}_{1},\mathcal{C}^{2}_{2},...\mathcal{C}^{2}_{J} \right\}$
            \State Flatten $\mathcal{C}^{2}$ where $\left| \mathcal{C}^{2} \right|=J^{'}\ge J$
            \State Update the sampling weight $W_{k}^{r}$ of each public data point $k$ (Eq. \eqref{eq:sample}, \eqref{eq:sample1})
            \EndIf
            \For{each $\mathcal{C}^{2}_{j^{'}}, j^{'}=1,2,...,J^{'}$ }
            \State $n_{j^{'}} \gets \sum_{i \in \mathcal{C}^{2}_{j^{'}}}n_{i}$
            \State $w_{\mathcal{L}_{j^{'}}^{s}}^{r} \gets \sum_{i \in \mathcal{C}^{2}_{j^{'}}}\frac{n_{i}}{n_{j^{'}}}w_{i}$
            \If{$H > h_{\text{th}}$}
            \State Update shared layers $\mathcal{L}_{j^{'}}^{s} \gets \mathcal{L}_{j^{'}}^{s} \times \psi^{r}$
            \EndIf
            \EndFor
    \EndFor
\State \textbf{Client} $i$ \textbf{executes:}
\State $w_{i}^{s} \gets w_{\mathcal{L}_{j}^{s}}^{r}$
\State $w_{i} \gets \left( w_{i}^{s},w_{i}^{per} \right)$
\For{each epoch $e=1,2,\ldots, E$}
\For{batch $\mathcal{D}_{B} \in \mathcal{D}_{i}$}
\State $w_{i} \gets w_{i}-\eta \nabla l\left( w_{i};\mathcal{D}_{B} \right)$
\EndFor
\EndFor
\State \textbf{Return} $w_{i}$ to the server
\end{algorithmic}
\end{algorithm}

At the beginning of the training procedure, the server initializes the global model parameters $w$ and defines the number of shared or personalized layers $\left[ \mathcal{L}_{s};\mathcal{L}_{per} \right]$. Meanwhile, all the clients are allocated to the same cluster $j$. And for each communication round $r$, we assume only $m=\text{max}\left( \delta \cdot K, 1 \right)$ clients are connected to the server, where $\delta$ is the connection ratio and $K$ is the total number of clients.

Then, the server sends clustered (shared) global model $w_{\mathcal{L}_{j}^{s}}^{r}$ ($w_{\mathcal{L}^{s}}$ for the first communication round) to all the clients $i \in [m]$ belonging to their respective cluster $j$. After that, each client $i$ upgrades the local shared model parameters $w_{i}^{s}$ using the received global model $w_{\mathcal{L}_{j}^{s}}^{r}$ and integrates it with the local personalization layer $w_{i}^{per}$ to form the local model $w_{i}$. Subsequently, $E$ epochs of training loops are performed on local batch data $\mathcal{D}_{B} \in \mathcal{D}_{i}$ and the updated model parameters $w_{i}$ will be returned back to the server (from line 36 to line 39 of Algorithm.\ref{algotirhm:fedtsd}).

Upon receiving $w_{i}$ from any connected client $i$, the server proceeds to sample $\mathcal{D}^{B}_{pub}$ from the public unlabeled dataset $\mathcal{D}_{pub}$. This sampled data is then utilized to compute the inference output $p_{i}$, as indicated in line 14 of Algorithm.\ref{algotirhm:fedtsd}. The resulting set of $m$ inference outputs, denoted as $I$, is subsequently employed to calculate the Hopkins statistic $H$. Once the Hopkins statistic $H$ exceeds the threshold $h_{\text{th}}$, the DBSCAN algorithm is executed using the similarity matrix $\text{Sim}$, which is calculated based on JS-divergence, as the clustering criterion The result of this clustering process is denoted as $\mathcal{C}^{1}$ and represents the first stage decoupling outcome.

The second stage of decoupling is applied to the results of $\mathcal{C}^{1}$ obtained in the previous stage. However, in contrast to the previous stage, the second stage employs model weight distance as the clustering metric (line 23-24 in Algorithm.\ref{algotirhm:fedtsd}). Consequently, each $\mathcal{C}^{1}_{j}$ may be further partitioned into several sub clusters, denoted as $\mathcal{C}^{2}_{j}$, which are then combined and flattened to form a newly generated cluster $\mathcal{C}^{2}$. By the way, the sampling weight $W^{r}_{k}$ of each public data point $k$ is updated by Eq. \eqref{eq:sample} and Eq. \eqref{eq:sample1}.

Afterwards, the server aggregates the client models within each cluster $\mathcal{C}_{j^{'}}^{2}, j^{'} \in \left[ J^{'} \right]$ in $\mathcal{C}^{2}$, where the aggregation weights are determined by the ratio of the local data size $n_{i}$ to total data size $n_{j^{'}}$ of all the clients in $\mathcal{C}_{j^{'}}^{2}$. It is worth noting that if the criterion of Hopkins statistic is not satisfied, then the model aggregation is performed based on the clustering outcome $\mathcal{C}^{2}$ from the last communication round $r-1$. Finally, the number of shared layers $\mathcal{L}_{j^{'}}^{s}$ is reduced by a dampening factor $\psi^{r}$ if two stages of decoupling is conducted. The aforementioned processes are repeated for $R$ communication rounds until convergence.

\section{Experiments}
To empirically verify the effectiveness and robustness of our proposed FedTSDP algorithm, extensive experimental studies are performed together with some state-of-the-art clustered FL methods on three image classification datasets. In this section, we first compare the learning performance of FedTSDP with other popular approaches on both IID and Non-IID data. Followed by the case study validating the effectiveness of Hopkins amended sampling strategy.

\subsection{Experimental Settings}
\subsubsection{Datasets}
Three image classification datasets are adopted in our simulations, namely CIFAR10 \cite{Krizhevsky09}, CIFAR100, and SVHN. As shown in Table.\ref{tab:dataset}, CIFAR10 contains 50000 training and 10000 testing $32 \times 32 \times 3$ images with 10 different kinds of objects, CIFAR100 contains 50000 training and 10000 testing $32 \times 32 \times 3$ images with 100 different types of objects, SVHN contains 73257 $32 \times 32 \times 3$ images with digits $0 \sim 9$.

\begin{table}[ht]
\caption{Description of the dataset}
\label{tab:dataset}
\begin{tabular}{llll}
    \hline
        Dataset & CIFAR10 & CIFAR100 & SVHN \\ 
    \hline
        Channel & 3 & 3 & 3 \\
        Class & 10 & 100 & 10 \\
        Size & 60000 & 60000 & 73257 \\ 
        Categories & objects & objects & digits \\
    \hline
\end{tabular}
\end{table}

To simulate the data distribution in FedTSDP, the original testing images are regarded as the public data $\mathcal{D}_{pub}$ on the server by removing their labels. And all training image data are evenly and randomly allocated to connected clients without overlap for IID experiments. While for Non-IID scenarios, each client is allocated a proportion of training data of each label class based on Dirichlet distribution $p_{c}\sim \text{Dir}_{k}\left( \beta \right)$, where $\beta$ is the concentration parameter and is selected to be $0.2$ and $0.5$ in our simulations. A smaller $\beta$ value would lead to a more unbalanced data partition. Except that, 20\% of the allocated data on each client are used for testing and the rest are used for training in all experiments.

\subsubsection{Models}
Two types of neural network models are selected as the global model used in our proposed FedTSDP. One is a convolutional neural network (CNN) with two convolutional layers:  one is a convolutional layer with a $3 \times 3$ kernel and 32 output channels, followed by another convolutional layer with a $3 \times 3$ kernel and 64 output channels. It further includes two fully connected layers, the first with 512 neurons and the second with 10 neurons. Additionally, two dropout layers are incorporated with dropout ratios of 0.25 and 0.5, respectively.

The other one is a more deeper ResNet18. It consists of one convolutional layer with a $3 \times 3$ kernel and 64 output channels, followed by a batch normalization layer \cite{pmlr-v37-ioffe15}. It is then followed by four BasicBlock structures, each of which comprises two groups of $3 \times 3$ convolutional layers and two batch normalization layers. Finally, fully connected layers are appended after the last BasicBlock.

\subsubsection{Algorithms Under Comparison}
Our proposed FedTSDP is compared to the following popular personalized and clustered FL algorithms: 

\begin{enumerate}
    \item \textbf{FedAvg\cite{fedavg}:} The earliest and most classical FL, which often serves as the baseline algorithm, adopts direct weighted averaging for model aggregation on the server.
    \item \textbf{FedProx\cite{fedprox}:} A influential modification of FedAvg, which incorporates local regularization terms to enhance convergence speed, is widely recognized as a baseline approach in personalized FL.
    \item \textbf{Ditto\cite{ditto}:} A recent and highly regarded work in personalized FL. Different from traditional model aggregation approaches, it adopts a unique strategy by directly replacing the global model with the local optimal model.
    \item \textbf{FedPer\cite{fedper}:} The pioneering work that introduces the concept of utilizing personalization layers to address data distribution heterogeneity in FL.
    \item \textbf{CFL\cite{cfl}:} An exemplary study that proposes bi-partitioning clustering for conducting multi-center FL, leading to successful personalization, is regarded as a notable contribution in this field.
\end{enumerate}

All hyperparameters settings for both our proposed FedTSDP and above five algorithms are kept consistent to ensure a fair comparison. Furthermore, it is crucial to acknowledge that our main objective is not to attain state-of-the-art performance but rather to obtain comparative outcomes. This consideration arises due to the sensitivity of FL to numerous hyperparameters that can significantly impact the results. 

\subsubsection{Other Settings}
All other experimental settings for FedTSDP are listed as follows:

\begin{itemize}
    \item Total number of clients $K$: 20
    \item Connection ratio $\delta$: 1.0
    \item Total number of communication rounds $R$: 200
    \item Number of local epochs: 2
    \item Training batch size: 50
    \item Local initial learning rate: 0.05
    \item Local learning momentum: 0.5
    \item Learning rate decay over each communication round: 0.95
    \item Epsilon threshold $\epsilon_{1}$, $\epsilon_{2}$ for DBSCAN: 0.15, 3.5
    \item Minimum points threshold for DBSCAN $\text{min}_{pts}$: 2
    \item Threshold of Hopkins statistic $h_{\text{th}}$: 0.65
    \item The dampening ratio of the shared layers $\psi^{r}$: 0.98
\end{itemize}

\subsection{Performance on IID data}
At first, the learning performance of FedTSDP on IID data are compared against five FL baseline algorithms. It is important to note that the learning performance is measured by the weighted average test accuracy of the global model (or within each cluster) on each local test set. In addition, for FL methods that incorporate personalization layers, it is sufficient to download and combine only the global shared layers with the local personalization layers for calculating the test accuracy.

\begin{table}[ht]
\caption{Final test performance on IID data}
\label{tab:iid}
\begin{tabular}{l|llllll}
    \hline
            \diagbox{Dataset}{Algorithm} & FedAvg & FedProx & Ditto & FedPer & CFL & FedTSDP \\
        \hline
            \multicolumn{7}{c}{CNN} \\
        \hline
            CIFAR10 & 71.79 & 72.37 & 72.01 & 66.14 & 69.95 & \textbf{72.98} \\ 
            CIFAR100 & 38.05 & 38.72 & 37.75 & 19.56 & 34.82 & \textbf{39.19} \\ 
            SVHN & 91.21 & 91.20 & 91.15 & 89.85 & 90.14 & \textbf{91.45} \\ 
        \hline
            \multicolumn{7}{c}{ResNet18} \\
        \hline
            CIFAR10 & 75.33 & 75.66 & 74.93 & 56.97 & 69.95 & \textbf{76.03} \\
            CIFAR100 & 38.45 & 38.89 & 38.18 & 18.16 & 34.82 & \textbf{39.74} \\
            SVHN & 93.68 & \textbf{93.82} & 93.80 & 88.76 & 90.14 & 93.72 \\
        \hline
\end{tabular}
\end{table}

\begin{figure}[ht]
  \includegraphics[width=0.92\textwidth]{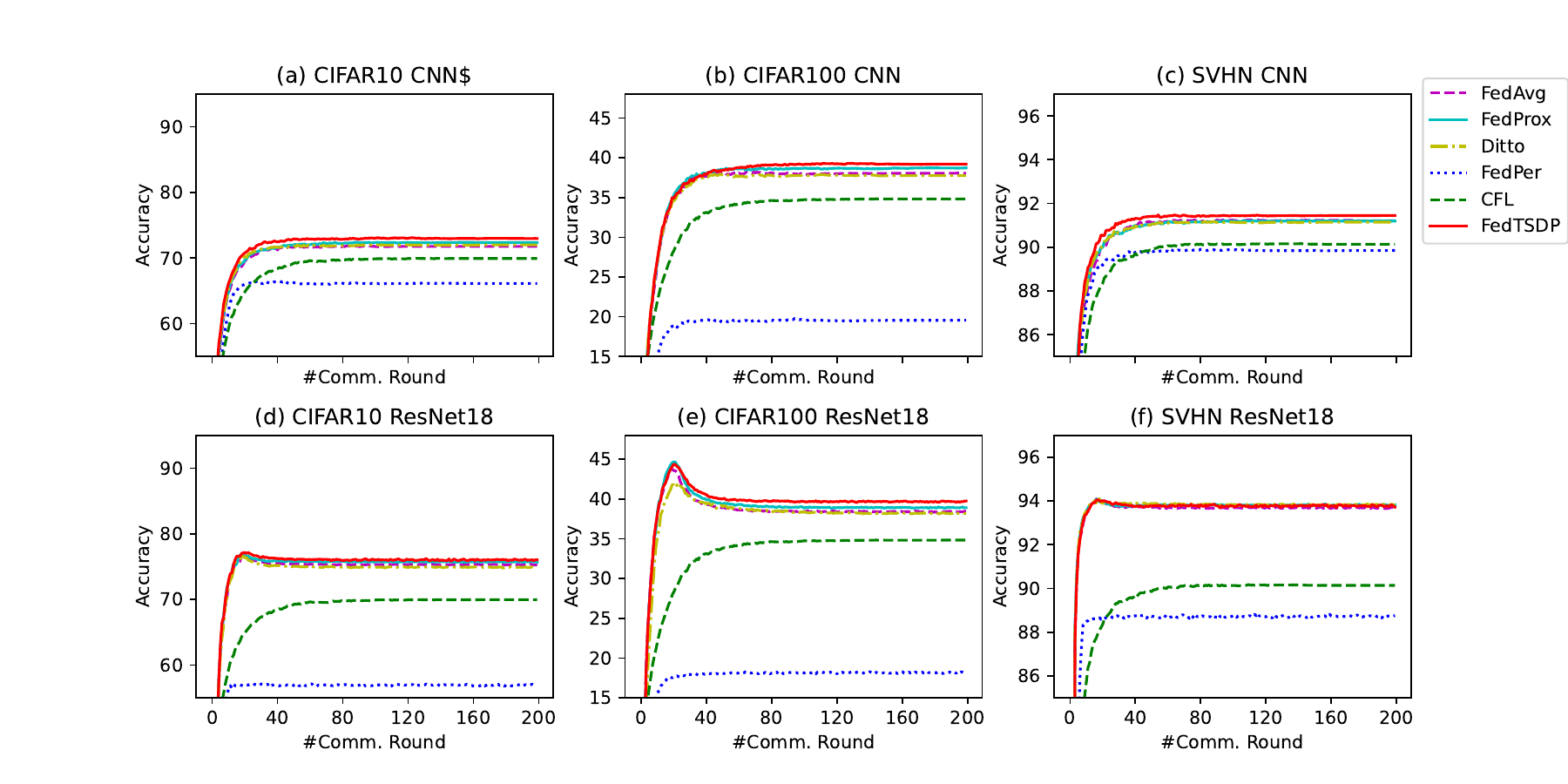}
\caption{The learning performance over communication rounds on IID data.}
\label{fig:iid}
\end{figure}

The final test results at the last communication round are presented in Table.\ref{tab:iid}. And it can be observed that FedTSDP demonstrates superior performance compared to other baseline algorithms on all three datasets. However, the differences in performance are not statistically significant due to homogeneous data distribution which may result in grouping all clients into one cluster. In this scenarios, the clustered FL essentially reverts to general FL. To be more specifically for CNN model, FedTSDP achieves the highest test accuracy of 72.98\%, 39.19\% and 91.45\% on CIFAR10, CIFAR100, and SVHN, respectively. These results are comparable to those obtained by FedAvg, FedProx, Ditto algorithms. Similar outcomes are observed as well, however, it is worth noting that FedProx achieves a test accuracy of 93.82\% on the SVHN dataset, which is slightly higher by 0.10\% compared to our proposed FedTSDP algorithm. It is surprising to see that the accuracy of FedPer is notably lower than that of all other algorithms. This discrepancy may be attributed to the fixed number of shared layers employed in FedPer, which restricts the ability of local personalization layers to learn information from other clients.

To take a closer look at the dynamic learning performance achieved by each algorithm, the convergence behavior across communication rounds are explored. As shown in Fig. \ref{fig:iid}, it is obvious that our proposed FedTSDP algorithm exhibits a slightly faster convergence speed compared to FedAvg, FedProx and Ditto. While CFL and FedPer has a relatively much slower convergence speed, which can be attributed to the 'strong' personalized operation that limits the assimilation of information from other participants. And in IID data scenarios, the incorporation of such information can enhance the learning process. Consequently, it can be concluded that clustering or personalization layers may not always be advantageous in the context of FL.

\subsection{Performance on Non-IID data}
\begin{table}[ht]
\caption{Final test accuracy on Non-IID data with Dirichlet distribution}
\label{tab:dir}
    \begin{tabular}{l|llllll}
    \hline
        \diagbox{Dataset}{Algorithm} & FedAvg & FedProx & Ditto & FedPer & CFL & FedTSDP \\ 
        \hline
            \multicolumn{7}{c}{CNN, $\beta=0.5$} \\
        \hline
            CIFAR10 & 69.66 & 69.66 & 70.75 & 78.47 & 66.06 & \textbf{79.29} \\ 
            CIFAR100 & 37.97 & 37.70 & 37.97 & 37.03 & 34.17 & \textbf{39.19} \\ 
            SVHN & 90.09 & 89.94 & 90.22 & 92.78 & 88.42 & \textbf{93.49} \\ 
        \hline
            \multicolumn{7}{c}{ResNet18, $\beta=0.5$} \\
        \hline
            CIFAR10 & 71.64 & 69.98 & 69.32 & 72.70 & 65.70 & \textbf{74.76} \\
            CIFAR100 & 38.94 & 38.75 & 38.51 & 34.16 & 38.18 & \textbf{43.17} \\
            SVHN & 92.92 & 93.05 & 93.44 & 92.99 & 92.71 & \textbf{95.31} \\
        \hline
            \multicolumn{7}{c}{CNN, $\beta=0.2$} \\
        \hline
            CIFAR10 & 67.35 & 67.87 & 66.96 & 82.59 & 73.84 & \textbf{84.75} \\
            CIFAR100 & 37.97 & 36.83 & 38.17 & 48.98 & 38.07 & \textbf{54.96} \\
            SVHN & 88.51 & 87.13 & 88.36 & 94.36 & 90.44 & \textbf{94.87} \\
        \hline
            \multicolumn{7}{c}{ResNet18, $\beta=0.2$} \\
        \hline
            CIFAR10 & 63.12 & 62.63 & 62.78 & 81.83 & 75.62 & \textbf{83.70} \\
            CIFAR100 & 38.62 & 39.22 & 37.75 & 45.99 & 39.25 & \textbf{46.69} \\
            SVHN & 92.51 & 91.85 & 91.46 & 95.08 & 92.19 & \textbf{95.57} \\
        \hline
    \end{tabular}
\end{table}

\begin{figure}[ht]
  \includegraphics[width=0.92\textwidth]{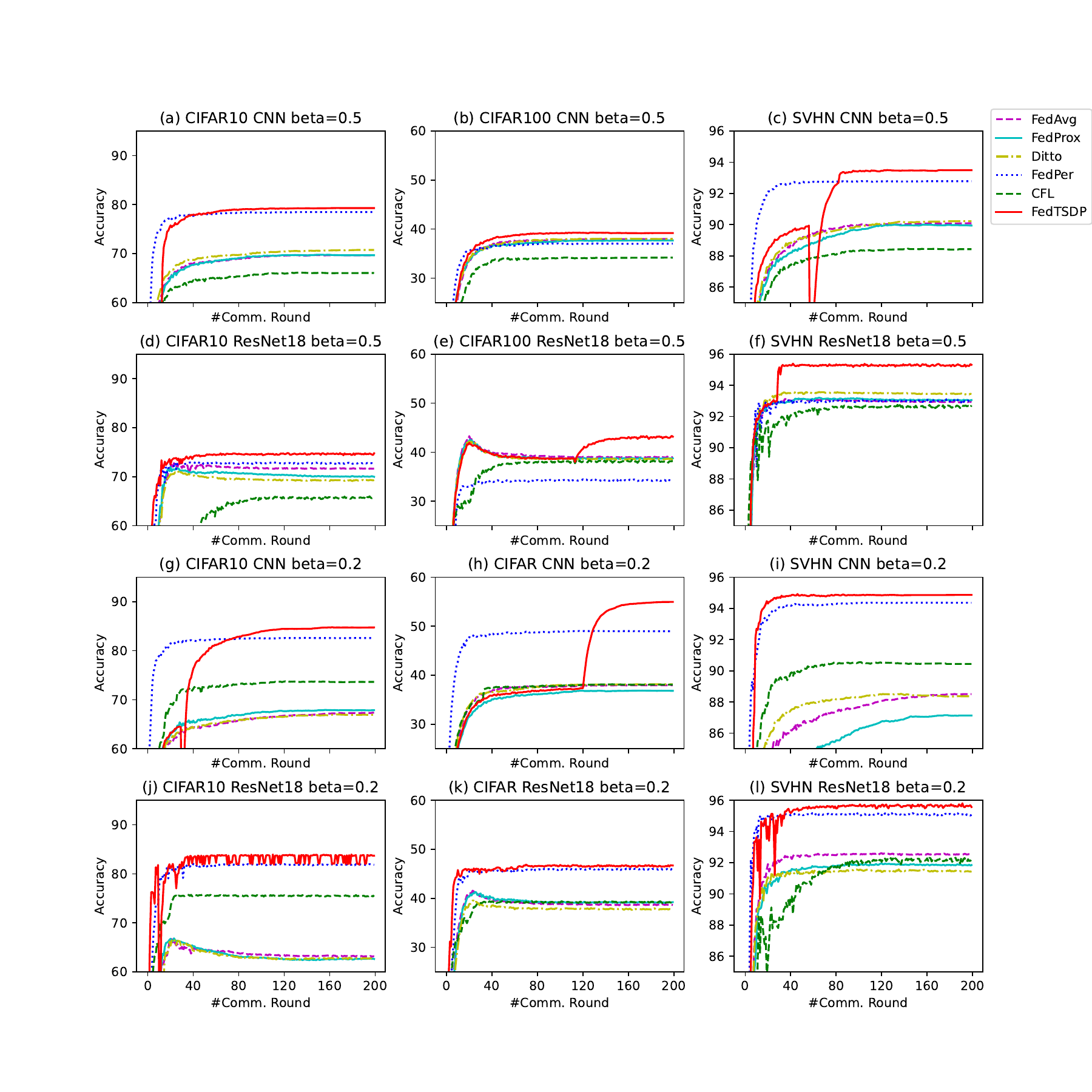}
\caption{The learning performance over communication rounds on Non-IID data partitioned follow Dirichlet distribution}
\label{fig:dir}
\end{figure}

To further assess the applicability of FedTSDP in more complex heterogeneous scenarios, we conducted evaluations of all the algorithms under various Non-IID settings. By varying the concentration parameter $\beta$ of the Dirichlet distribution ($\beta=0.2, 0.5$ in our experiments), it is possible to generate Non-IID data with varying degrees of data skew. Furthermore, the data partition method used in FedAvg is also utilized in our simulations, where each client owns data samples with a fixed number of two label classes. The client data generated using this method is often regarded as extreme Non-IID data due to the significant variations in data distributions among the clients.

The results on both CNN and ResNet18 using Dirichlet partition method are shown in Table.\ref{tab:dir}. It is evident that our proposed FedTSDP outperforms other baseline algorithms in scenarios involving Non-IID data with varying degrees of data heterogeneity. In contrast to the results obtained on IID data, FedPer consistently achieves the second highest test accuracy, surpassing the FedAvg algorithm by approximately 9\% on CNN model. This performance gap becomes more pronounced in non-IID scenarios, particularly with a concentration parameter of $\beta=0.2$, where FedPer achieves a remarkable 15\% higher test accuracy compared to other algorithms. In addition to FedPer, the CFL algorithm also demonstrates promising results in handling heterogeneous data distributions, particularly in scenarios with a concentration parameter of $\beta=0.2$, In such cases, CFL achieves an approximately 5\% higher test accuracy compared to non-clustered FL algorithms. These phenomena implicitly highlight the advantages of both clustering and incorporating personalization layers in Non-IID scenarios, underscoring the importance of integrating these approaches together.

Regarding the final results on CIFAR10 dataset (Table. \ref{tab:2shards}), where each client is assigned only two classes of objects, FedTSDP demonstrates superior learning performance. Specifically, it achieves a test accuracy of 90.49\% on the CNN model and 90.20\% on the ResNet18 model. These accuracies are 0.29\% and 0.03\% higher, respectively, compared to FedPer. FedAvg, FedProx, and Ditto share similar learning performance with a test accuracy of approximately 63\%, which is about 27\% lower than FedTSDP and FedPer. This empirically prove the effectiveness of personalization layers to deal with highly Non-IID data in FL.

\begin{table}[ht]
\caption{Final test accuracy on CIFAR10 dataset using extreme Non-IID partition method}
\label{tab:2shards}
    \begin{tabular}{l|llllll}
    \hline
            \diagbox{Model}{Algorithm} & FedAvg & FedProx & FedPer & Ditto & CFL & FedTSDP \\
        \hline
            CNN & 62.88 & 63.63 & 90.22 & 63.59 & 75.63 & \textbf{90.49} \\
            ResNet18 & 54.31 & 50.84 & 90.17 & 55.57 & 78.80 & \textbf{90.20} \\
        \hline
    \end{tabular}
\end{table}

Similarly, the real-time test performance over communication rounds are illustrated in Fig. \ref{fig:dir}, and the following three observations can be made. First, our proposed FedTSDP algorithm in general converges faster than other baseline algorithms. Second, there may be instances of \emph{sharp increases or decreases} observed prior to the midpoint of the communication rounds, which is typically attributed to sudden clusters variations. It is important to note that FedTSDP does not perform clustering unless the Hopkins statistic criterion is satisfied. Thus, after an extended period of unchanged clustered training, aggregating re-clustered client models may cause a sudden performance drop. In addition, the sudden performance degradation also comes from adaptive layer adjustment. Decaying the number of shared layers directly changes the scale of local personalization layers which may bring in unexpected learning bias. And each modified personalization layers require several communication rounds of local training for recovery. Third, the performance gain on highly Non-IID data mainly comes from personalization layers, as FedPer converges even faster than FedTSDP at the beginning of the training period.

What is worth mentioning is that well-organized groupings of clusters and appropriately shared layers are expected to accelerate the convergence speed of a FL system. This may lead to an improvement boost in model performance after a few rounds of training recovery, which further substantiates the effectiveness and significance of our proposed FedTSDP.

By the way, it is interesting to observe that the performance of our algorithm on SVHN dataset fluctuates, showing opposite behaviors on CNN and ResNet18. This is because SVHN dataset is comparatively much easier to be trained compared with other datasets. The learning model tends to converge during the early stages of the federated training period, after which it may begin to exhibit fluctuations. ResNet18, as a deeper and more complex architecture compared to a simple CNN, is more likely to achieve faster convergence. We also repeat the experiments several times, and the means and variances are reported in Table. \ref{tab:svhn}.

\begin{table}[ht]
\caption{Final test accuracy mean/variance on SVHN dataset}
\label{tab:svhn}
    \begin{tabular}{l|lll}
    \hline
            \diagbox{Model}{Algorithm} & FedAvg & FedProx & FedPer \\
        \hline
            CNN & 90.36\%$\pm$0.04 & 90.12\%$\pm$0.04 & 92.85\%$\pm$0.01 \\
            ResNet18 & 93.03\%$\pm$0.03 & 93.11\%$\pm$0.00 & 93.04\%$\pm$0.00 \\
        \hline
            \diagbox{Model}{Algorithm} & Ditto & CFL & FedTSDP \\
        \hline
            CNN & 90.18\%$\pm$0.00 & 89.42\%$\pm$0.51 & \textbf{93.10\%$\pm$0.11} \\
            ResNet18 & 93.10\%$\pm$0.11 & 92.39\%$\pm$0.09 & \textbf{93.69\%$\pm$0.04} \\
    \hline
    \end{tabular}
\end{table}

\subsection{Ablation Study on Two Stage Decoupling}
To testify the validation of proposed two-stage decoupling scheme, we decompose FedTSDP and elaborate each stage one by one. For simplicity, this ablation study is conducted on CIFAR10 dataset with $\beta=0.5$.

\begin{figure}[ht]
  \includegraphics[width=0.92\textwidth]{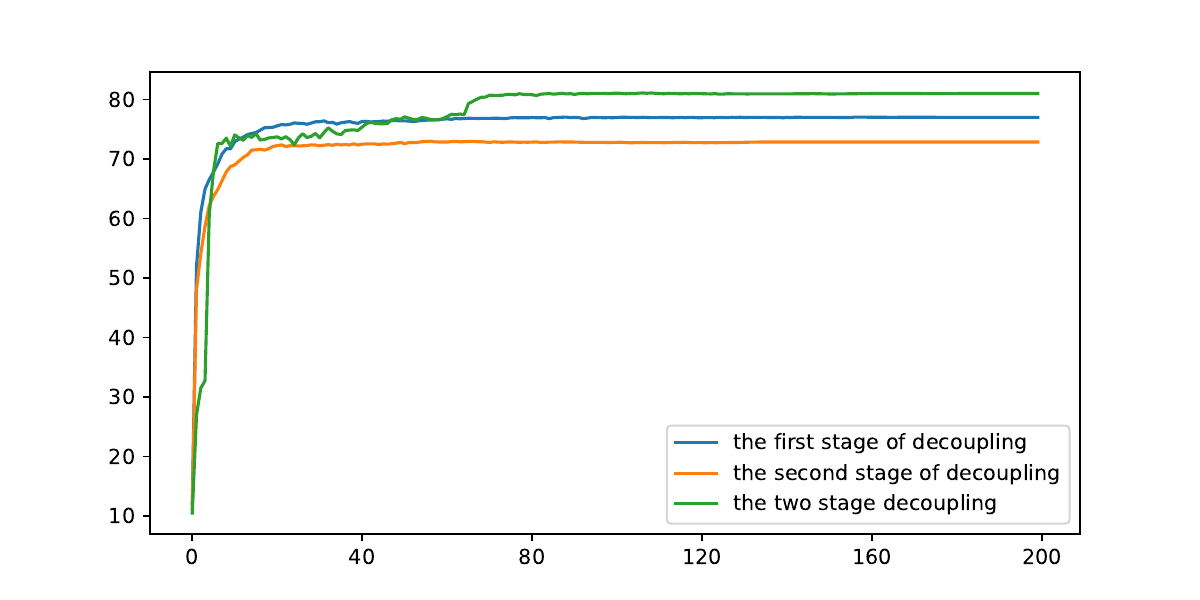}
\caption{Ablation study on two-stage decoupling, where both single stage and two-stage decoupling are included.}
\label{fig:stage}
\end{figure}

The corresponding outcomes over communication rounds are shown in Fig. \ref{fig:stage}, where the learning curves of the first stage of decoupling, the second stage of decoupling and two-stage decoupling are included, respectively. Overall, it is apparent to see that two-stage decoupling scheme converges slightly slower than the first stage of decoupling at the early steps of the period, while it achieves the best learning performance after approximate 60 communication round. And the second stage of decoupling shows the worst convergence property among these three methods. This empirically illustrates the insufficiency of single stage decoupling, and substantiates the effectiveness and significance of our proposed two-stage decoupling approach that can not only reflect the local data distribution, but also preserve the statistical characteristics of local models, especially for the context of Non-IID data.

\subsection{Analysis on Hopkins Amended Sampling}
To further validate the effectiveness of the proposed Hopkins amended sampling strategy, we conduct a dedicated experiment using both IID and non-IID data. For a brevity sake, this study is only performed on CNN model with SVHN dataset.

\begin{figure}[ht]
  \includegraphics[width=0.8\textwidth]{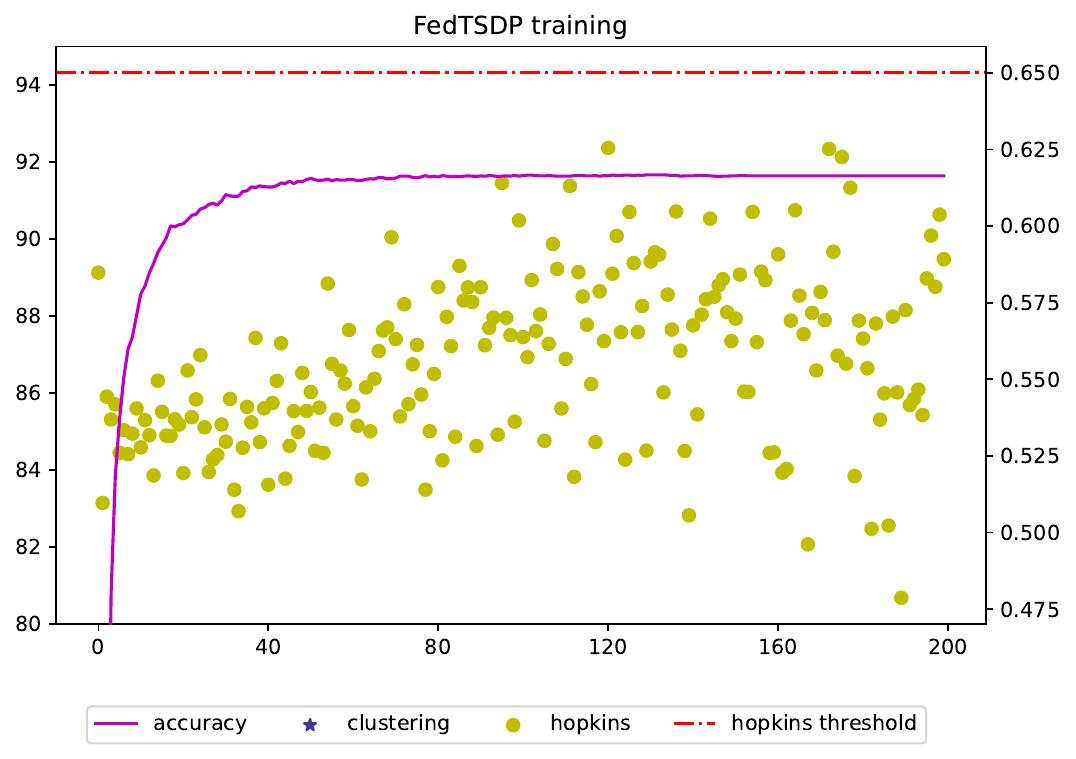}
\caption{Hopkins statistic on IID data. The left axis represents the accuracy, while the right axis corresponds to the Hopkins statistic. It can be observed that in each round, none of the Hopkins statistics reach the threshold value.}
\label{fig:hopkins_iid}
\end{figure}

\begin{figure}[ht]
  \includegraphics[width=0.8\textwidth]{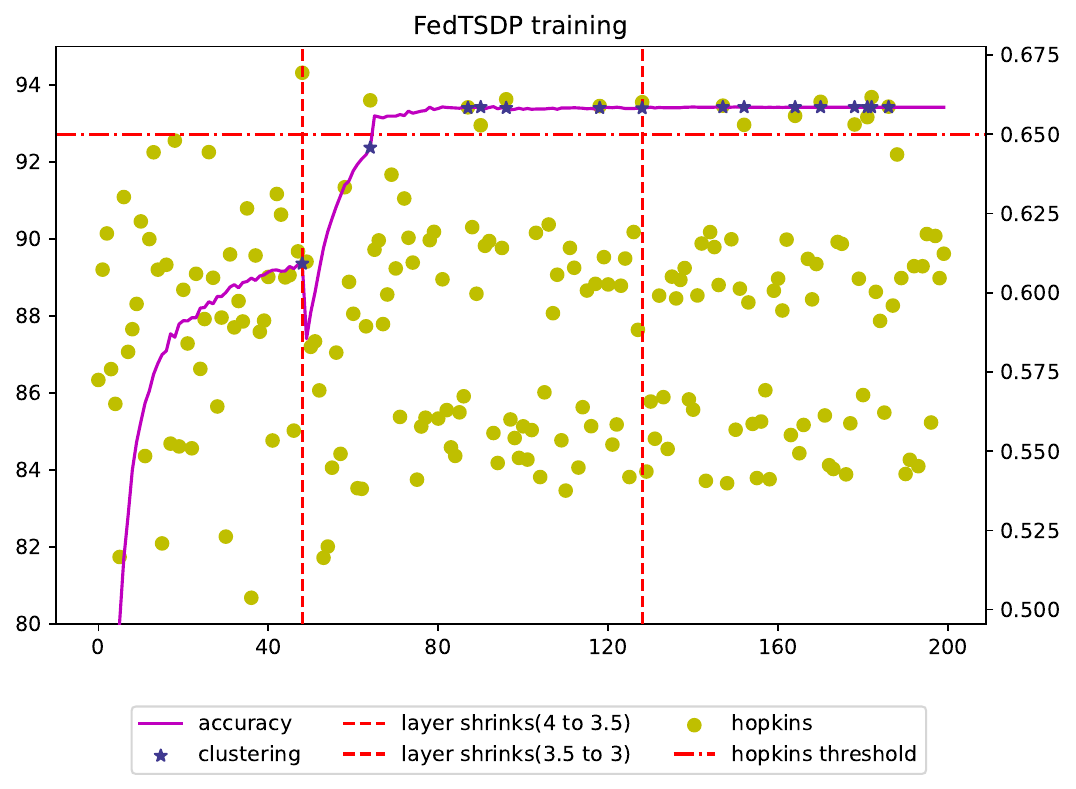}
\caption{The Hopkins statistic on Non-IID data. The two dashed vertical lines indicate the occurrence of layer shrinking.}
\label{fig:hopkins_noniid}
\end{figure}

In the IID results illustrated in Fig. \ref{fig:hopkins_iid}, the values on the left axis represent the test accuracy, while the values on the right axis represent Hopkins statistic. Furthermore, the dot points represent the calculated Hopkins statistic and the solid curve is test accuracy over communication rounds. It is evident that no clustering occurs as observed from the plot. As the training process progresses, the Hopkins statistic gradually increases, indicating a potential increase in model divergence. However, it does not exceed the Hopkins threshold (the dashed horizontal line), indicating that clustering does not occur under IID FL environment.

The non-IID situation is depicted in Fig. \ref{fig:hopkins_noniid}. As observed, the Hopkins statistic exhibits significant fluctuations, indicating a high degree of client bias. Additionally, the upward trend of the Hopkins statistic is attributed to the adopted sampling strategy. Except that, it is intriguing to observe a noticeable performance degradation occurring around the 50th communication round, coinciding with the reduction of shared layers from 4 to 3.5 and the simultaneous occurrence of clustering.

\section{Conclusion}
In this paper, we introduce FedTSDP, a two-stage decoupling mechanism that incorporates adaptive personalization layers to effectively handle heterogeneity issues arising from varying degrees of data skew. FedTSDP leverages unsupervised clustering using unlabeled data on the server, along with the Hopkins amended sampling technique to maintain active and valuable data information. Furthermore, it employs a dynamic adjustment strategy for the shared layers to automatically handle both IID and Non-IID data.

Extensive experiments are performed to compare the proposed FedTSDP with five baseline FL algorithms. The results demonstrate that the solutions obtained by our method exhibit superior or comparable learning performance, and are well suited to varying levels of data heterogeneity. This advantage can be attributed to the Hopkins statistic, which selectively performs clustering only when there is a high clustering tendency present in the client data. In addition, it is intriguing to discover that personalization layers in FL may not always confer learning performance benefits, especially when dealing with homogeneous data. Overall, our proposed FedTSDP algorithm showcases promising performance on both heterogeneous and homogeneous data distributions.

The present work is an important initial step towards incorporating hierarchical clustering and personalization layers in FL. Despite the encouraging empirical results we have obtained, it is important to note that the classification performance of models over communication rounds may occur a sharp drop. Therefore, in the future, our focus will be on developing more robust FL algorithm to make the learning progress more stable.

\begin{acknowledgements}
This work was supported in part by the National Science Foundation of China (NSFC) under Grant 62272201, and in part by the Wuxi Science and Technology Development Fund Project under Grant K20231012.
\end{acknowledgements}

\section*{Declarations}
On behalf of all authors, the corresponding author states that there is no conflict of interest.

\bibliographystyle{spmpsci}      
\bibliography{ref}   

\end{document}